%% file: naaclhlt2019.tex
\documentclass[11pt,a4paper]{article}
\usepackage[hyperref]{naaclhlt2019}
\usepackage{times}
\usepackage{latexsym}

\usepackage{url}

\usepackage{amsmath}
\usepackage[export]{adjustbox}
\usepackage{graphicx}
\usepackage{multirow}
\usepackage{hyperref}
\usepackage{dirtytalk}
\usepackage{subcaption}

\aclfinalcopy 

\title{Knowledge-Augmented Language Model and Its Application to Unsupervised Named-Entity Recognition}

\author{Angli Liu \\
  Facebook AI \\
  {\tt anglil@cs.washington.edu} \\\And
  Jingfei Du \\
  Facebook AI \\
  {\tt jingfeidu@fb.com} \\\And
  Veselin Stoyanov \\
  Facebook AI \\
  {\tt ves@fb.com} \\}

\date{}

\begin{document}
\maketitle
\begin{abstract}
  Traditional language models are unable to efficiently model entity names observed in text. All but the most popular named entities appear infrequently in text providing insufficient context. Recent efforts have recognized that context can be generalized between entity names that share the same type (e.g., \emph{person} or \emph{location}) and have equipped language models with access to an external knowledge base (KB). Our Knowledge-Augmented Language Model (KALM) continues this line of work by augmenting a traditional model with a KB. Unlike previous methods, however, we train with an end-to-end predictive objective optimizing the perplexity of text. We do not require any additional information such as named entity tags. In addition to improving language modeling performance, KALM learns to recognize named entities in an entirely unsupervised way by using entity type information latent in the model. On a Named Entity Recognition (NER) task, KALM achieves performance comparable with state-of-the-art supervised models. Our work demonstrates that named entities (and possibly other types of world knowledge) can be modeled successfully using predictive learning and training on large corpora of text without any additional information.
\end{abstract}

\input{intro.tex}
\input{related_work.tex}
\input{model.tex}

\input{ner.tex}
\input{experiments.tex}
\input{conclusion.tex}
\clearpage



\bibliography{naaclhlt2019}
\bibliographystyle{acl_natbib}

\end{document}

%% file: intro.tex
\section{Introduction}
\label{intro}
Language modeling is a form of unsupervised learning that allows language properties to be learned from large amounts of unlabeled text. As components, language models are useful for many Natural Language Processing (NLP) tasks such as generation \citep{parvez2018building} and machine translation \citep{bahdanau2014neural}. Additionally, the form of predictive learning that language modeling uses is useful to acquire text representations that can be used successfully to improve a number of downstream NLP tasks \cite{peters2018deep, devlin2018bert}. In fact, models pre-trained with a predictive objective have provided a new state-of-the-art by a large margin.

Current language models are unable to encode and decode factual knowledge such as the information about entities and their relations. Names of entities are an open class. While classes of named entities (e.g., \emph{person} or \emph{location}) occur frequently, each individual name (e.g, \emph{Atherton} or \emph{Zhouzhuang}) may be observed infrequently even in a very large corpus of text. As a result, language models learn to represent accurately only the most popular named entities. In the presence of external knowledge about named entities, language models should be able to learn to generalize across entity classes. For example, knowing that \emph{Alice} is a name used to refer to a person should give ample information about the context in which the word may occur (e.g., \emph{Bob visited \underline{Alice}}).

In this work, we propose Knowledge Augmented Language Model (KALM), a language model with access to information available in a KB. Unlike previous work, we make no assumptions about the availability of additional components (such as Named Entity Taggers) or annotations. Instead, we enhance a traditional LM with a gating mechanism that controls whether a particular word is modeled as a general word or as a reference to an entity. We train the model end-to-end with only the traditional predictive language modeling perplexity objective. As a result, our system can model named entities in text more accurately as demonstrated by reduced perplexities compared to traditional LM baselines. In addition, KALM learns to recognize named entities completely unsupervised by interpreting the predictions of the gating mechanism at test time. In fact, KALM learns an unsupervised named entity tagger that rivals in accuracy supervised counterparts.

KALM works by providing a language model with the option to generate words from a set of entities from a database. An individual word can either come from a general word dictionary as in traditional language model or be generated as a name of an entity from a database. Entities in the database are partitioned by type. The decision of whether the word is a general term or a named entity from a given type is controlled by a gating mechanism conditioned on the context observed so far. Thus, KALM learns to predict whether the context observed is indicative of a named entity of a given type and what tokens are likely to be entities of a given type.

The gating mechanism at the core of KALM is similar to attention in Neural Machine Translation \citep{bahdanau2014neural}. As in translation, the gating mechanism allows the LM to represent additional latent information that is useful for the end task of modeling language. The gating mechanism (in our case entity type prediction) is latent and learned in an end-to-end manner to maximize the probability of observed text. Experiments with named entity recognition show that the latent mechanism learns the information that we expect while LM experiments show that it is beneficial for the overall language modeling task. 

This paper makes the following contributions:
\begin{itemize}
    \item Our model, KALM, achieves a new state-of-the art for Language Modeling on several benchmarks as measured by perplexity.
    \item We learn a named entity recognizer without any explicit supervision by using only plain text. Our unsupervised named entity recognizer achieves a performance on par with the state-of-the supervised methods.
    \item We demonstrate that predictive learning combined with a gating mechanism can be utilized efficiently for generative training of deep learning systems beyond representation pre-training. 
\end{itemize}

%% file: related_work.tex
\section{Related Work}
\label{relatedwork}
Our work draws inspiration from \citet{ahn2016neural}, who propose to predict whether the word to generate has an underlying fact or not. Their model can generate knowledge-related words by copying from the description of the predicted fact. While theoretically interesting, their model functions only in a very constrained setting as it requires extra information: a shortlist of candidate entities that are mentioned in the text.

Several efforts successfully extend LMs with entities from a knowledge base and their types, but require that entity models are trained separately from supervised entity labels. \citet{parvez2018building} and \citet{xin2018put} explicitly model the type of the next word in addition to the word itself. In particular, \citet{parvez2018building} use two LSTM-based language models, an entity type model and an entity composite (entity type) model. \citet{xin2018put} use a similarly purposed entity typing module and a LM-enhancement module. 
Instead of entity type generation, \citet{gu2018language} propose to explicitly decompose word generation into sememe (a semantic language unit of meaning) generation and sense generation, but requires sememe labels.
\citet{yang2016reference} propose a pointer-network LM that can point to a 1-D or 2-D database record during inference. At each time step, the model decides whether to point to the database or the general vocabulary. 


Unsupervised predictive learning has been proven effective in improving text understanding. 
ELMo \citep{peters2018deep} and BERT \citep{devlin2018bert} used different unsupervised objectives to pre-train text models which have advanced the state-of-the-art for many NLP tasks. 
Similar to these approaches KALM is trained end-to-end using a predictive objective on large corpus of text.

Most unsupervised NER models are rule-based \cite{collins1999unsupervised, etzioni2005unsupervised, nadeau2006unsupervised} and require feature engineering or parallel corpora \citep{munro2012accurate}. \citet{yang2017leveraging} incorporate a KB to the CRF-biLSTM model \citep{lample2016neural} by embedding triples from a KB obtained using \textit{TransE} \citep{bordes2013translating}. \citet{peters2017semi} add pre-trained language model embeddings as knowledge to the input of a CRF-biLSTM model, while still requiring labels in training. And \citet{zhou2018zero} are concerned only with zero-shot NER inference.

To the best of our knowledge, KALM is the first unsupervised neural NER approach. As we discuss in Section \ref{sec:ner}, KALM achieves results comparable to supervised CRF-biLSTM models.

%% file: model.tex
\section{Knowledge-Augmented Language Model}
\label{model}
KALM extends a traditional, RNN-based neural LM.
As in traditional LM, KALM predicts probabilities of words from a vocabulary $V_g$, but it can also generate words that are names of entities of a specific type. Each entity type has a separate vocabulary $\{V_1,...,V_{K}\}$ collected from a KB. KALM learns to predict from context whether to expect an entity from a given type and generalizes over entity types.

\subsection{RNN language model}
At its core, a language model predicts a distribution for a word $y_{t+1}$ given previously observed words $c_t:=[y_{1},..., y_{t-1}, y_{t}]$. Models are trained by maximizing the likelihood of the observed next word.
In an LSTM LM, the probability of a word, $P(y_{t+1}|c_t)$, is modeled from the hidden state of an LSTM \citep{hochreiter1997long}:
\begin{align}
    P(y_{t+1}=i|c_t) = \frac{\exp(\boldsymbol{W}^p_{i,:}\cdot\boldsymbol{h}_t)}{\sum\limits_{w=1}^{|V_g|} \exp(\boldsymbol{W}^p_{w,:}\cdot\boldsymbol{h}_t)}
    \label{eq:lm_obj}
\end{align}
\begin{align}
    \boldsymbol{h}_t, \boldsymbol\gamma_t = lstm(\boldsymbol{h}_{t-1}, \boldsymbol\gamma_{t-1}, \boldsymbol{y}_t)
    \label{eq:lstm}
\end{align}
where $lstm$ refers to the LSTM step function and $h_i$, $\boldsymbol\gamma_{i}$ and $y_i$ are the hidden, memory and input vectors, respectively. $\boldsymbol{W}^p$ is a projection layer that converts LSTM hidden states into logits that have the size of the vocabulary $|V_g|$.

\subsection{Knowledge-Augmented Language Model}
KALM builds upon the LSTM LM by adding type-specific entity vocabularies $V_1, V_2, ... , V_{K}$ in addition to the general vocabulary $V_g$. Type vocabularies are extracted from the entities of specific type in a KB. For a given word, KALM computes a probability that the word represents an entity of that type by using a type-specific projection matrix $\{\boldsymbol{W}^{p,j}|j=0,...,K\}$. The model also computes the probability that the next word represents different entity types given the context observed so far. The overall probability of a word is given by the weighted sum of the type probabilities and the probability of the word under the give type. 


More precisely, let $\tau_{i}$ be a latent variable denoting the type of word $i$. We decompose the probability in Equation \ref{eq:lm_obj} using the type $\tau_{t+1}$: 

\begin{align}
    P(y_{t+1}|c_t) &= \sum_{j=0}^{K}& P(y_{t+1}, \tau_{t+1}=j|c_t) \nonumber\\
    &= \sum_{j=0}^{K}& P(y_{t+1} | \tau_{t+1}=j, c_t) \nonumber \\
        &&\cdot P(\tau_{t+1}=j | c_t) \label{eq:cond_prob}
\end{align}

Where $P(y_{t+1} | \tau_{t+1}, c_t)$ is a distribution of entity words of type $\tau_{t+1}$. As in a general LM, it is computed by projecting the hidden state of the LSTM and normalizing through softmax (eq. \ref{eq:tok_cond_prob}). The type-specific projection matrix $\boldsymbol{W}^{p,j}$ is learned during training.

\begin{figure}
    \centering
    \begin{subfigure}[b]{0.48\textwidth}
        \centering
        \includegraphics[width=\textwidth]{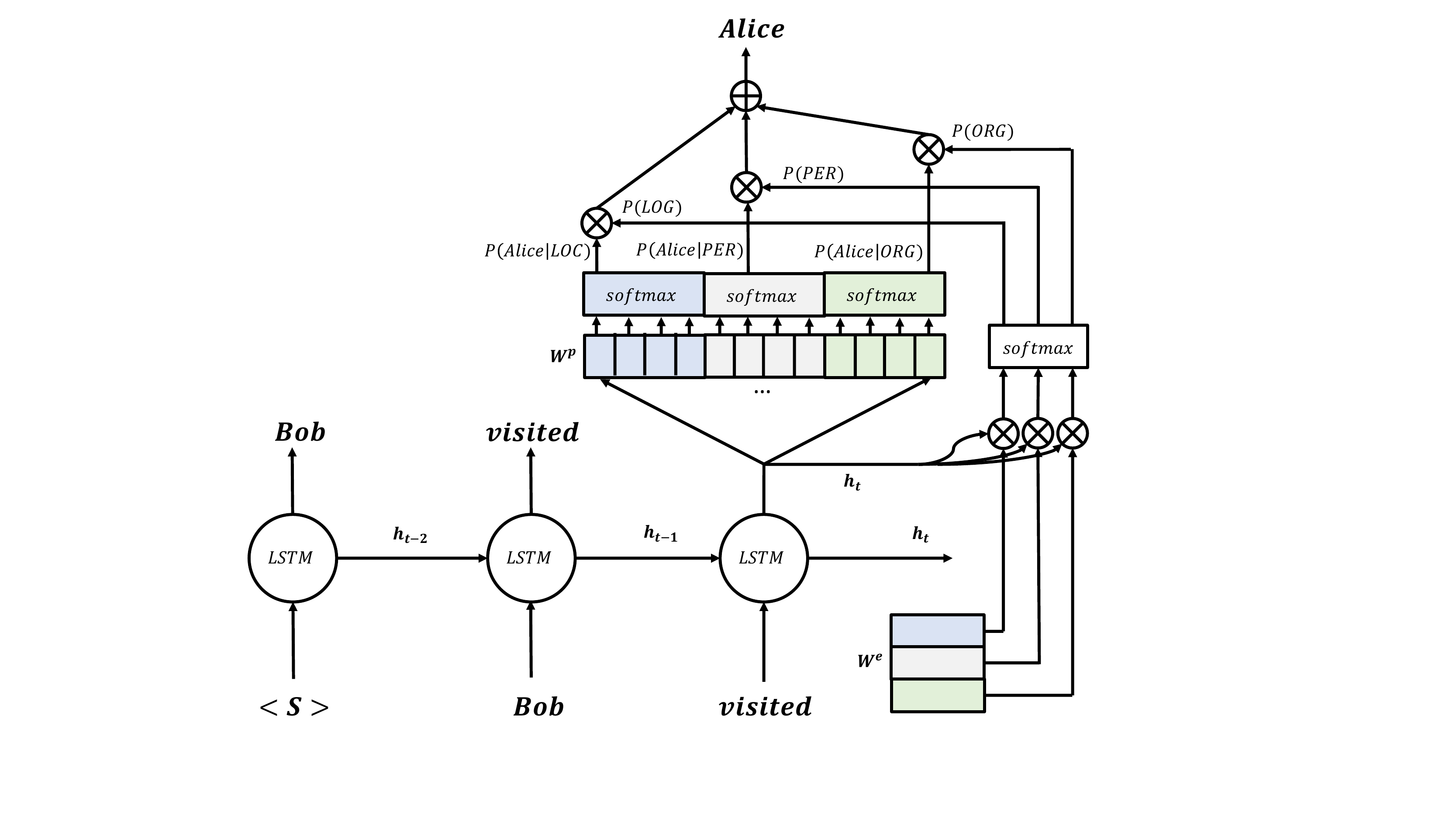}
        \caption{Basic model architecture of KALM.}
        \label{model_illustration}
    \end{subfigure}
    ~
    \begin{subfigure}[b]{0.48\textwidth}
        \centering
        \includegraphics[width=\textwidth]{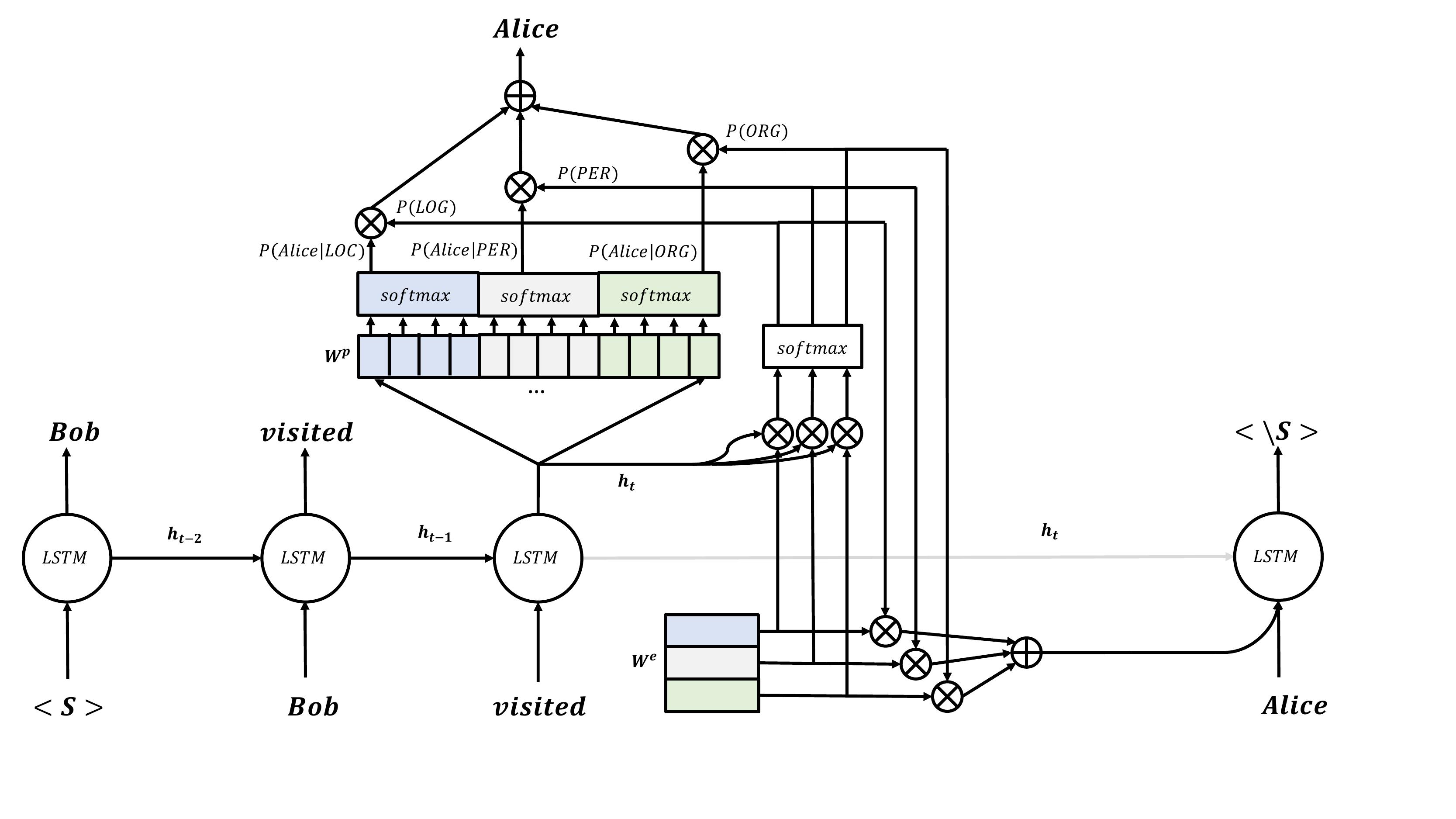}
        \caption{Adding type representation as input to KALM.}
        \label{type_feature}
    \end{subfigure}
    \caption{KALM's architectures}
\end{figure}

We maintain a type embedding matrix $\boldsymbol{W}^{e}$ and use it in a similar manner to compute the probability that the next word has a given type $P(\tau_{t+1} | c_t)$ (eq. \ref{eq:type_prob}). The only difference is that we use an extra projection matrix, $\boldsymbol{W}^h$ to project $\boldsymbol{h}_t$ into lower dimensions.
Figure \ref{model_illustration} illustrates visually the architecture of KALM. 

\begin{align}
    P(y_{t+1}=i | \tau_{t+1}=j, c_t) = \frac{\exp(\boldsymbol{W}^{p,j}_{i,:} \cdot \boldsymbol{h}_t)}{\sum\limits_{w=1}^{|V_j|} \exp(\boldsymbol{W}^{p,j}_{w,:} \cdot \boldsymbol{h}_t)} \label{eq:tok_cond_prob}
\end{align}
\begin{align}
    P(\tau_{t+1}=j | c_t) = \frac{\exp(\boldsymbol{W}^{e}_{j,:}\cdot(\boldsymbol{W}^h\cdot\boldsymbol{h}_t))}{\sum\limits_{k=0}^{K}\exp(\boldsymbol{W}^{e}_{k,:}\cdot(\boldsymbol{W}^h\cdot\boldsymbol{h}_t))} \label{eq:type_prob}
\end{align}





\subsection{Type representation as input}

In the base KALM model the input for word $y_t$ consists of its embedding vector $\boldsymbol{y}_t$. We enhance the base model by adding as inputs the embedding of the type of the previous word. As type information is latent, we represent it as the weighted sum of the type embeddings weighted by the predicted probabilities:
\begin{align}
    \boldsymbol{\nu}_{t+1} = \sum_{j=0}^{K} P(\tau_{t+1}=j|c_t) \cdot \boldsymbol{W}^{e}_{j,:}  \label{eq:inp_feedfwd}
\end{align}
\begin{align}
    \tilde{\boldsymbol{y}}_{t+1} = [\boldsymbol{y}_{t+1}; \boldsymbol\nu_{t+1}] \label{eq:concat_type}
\end{align}

\noindent
$P(\tau_{t+1}=j|c_t)$ is computed using Equation \ref{eq:type_prob} and $e_j$ is the type embedding vector.

Adding type information as input serves two purposes: in the forward direction, it allows KALM to model context more precisely based on predicted entity types. During back propagation, it allows us to learn latent types more accurately based on subsequent context. The model enhanced with type input is illustrated in Figure \ref{type_feature}.


%% file: ner.tex
\section{Unsupervised NER}
\label{app_ner}

The type distribution that KALM learns is latent, but we can output it at test time and use it to predict whether a given word refers to an entity or a general word. We compute
 $P(\tau_{t+1} | c_t)$ using eq. \ref{eq:type_prob} and use the most likely entity type as the named entity tag for the corresponding word $y_{t+1}$. 

This straightforward approach, however, predicts the type based solely on the left context of the tag being predicted. In the following two subsections, we discuss extensions to KALM that allow it to utilize the right context and the word being predicted itself.


\subsection{Bidirectional LM}
\label{bidirectional}
While we cannot use a bidirectional LSTM for generation, we can use one for NER, since the entire sentence is known. 

For each word, KALM generates the hidden vectors $\boldsymbol{h}_{l,t}$ and $\boldsymbol{h}_{r,t}$ representing context coming from left and right directions, as shown in Equations \ref{eq:bi_l} and \ref{eq:bi_r}.


\begin{align}
    \boldsymbol{h}_{l,t}, \boldsymbol{\gamma}_{l,t} &= lstm_l(\boldsymbol{h}_{l,t-1}, \boldsymbol{\gamma}_{l,t-1}, [\boldsymbol{y}_{l,t}; \boldsymbol{\nu}_{l,t}]) \label{eq:bi_l}\\
    \boldsymbol{h}_{r,t}, \boldsymbol{\gamma}_{r,t} &= lstm_r(\boldsymbol{h}_{r,t+1}, \boldsymbol{\gamma}_{r,t+1}, [\boldsymbol{y}_{r,t}; \boldsymbol{\nu}_{r,t}]) \label{eq:bi_r}
\end{align}

We concatenate the hidden vectors from the two directions to form an overall context vector $\boldsymbol{h}_{t}$, and generate the final type distribution using Equation \ref{eq:type_prob}.

Training the bidirectional model requires that we initialize the hidden and cell states from both ends of a sentence. Suppose the length of a sentence is $n$. The the cross entropy loss is computed for only the $n-2$ symbols in the middle. Similarly, we compute only the types of the $n-2$ symbols in the middle during inference.

\subsection{Current word information}
\label{use_prior}
Even bidirectional context is insufficient to predict the word type by itself. Consider the following example: 
\textit{Our computer models indicate Edouard is going to $\_\_\_$}\footnote{All the examples are selected from the CoNLL 2003 training set.}
The missing word can be either a location (e.g., \textit{London}), or a general word (e.g., \textit{quit}). In an NER task we observe the underlined words:
\textit{Our computer models indicate Edouard is going to \underline{London}.}
In order to learn predictively, we cannot base the type prediction on the current token. Instead, we can use a prior type information  $P(\tau_t|y_t)$, pre-computed from entity popularity information available in many KBs. We incorporate the prior information $P(\tau_t|y_t)$ in two different ways described in the two following subsections.

\subsubsection{Decoding with type prior} 
\label{prior_in_dec}
We incorporate the type prior $P(\tau_t|y_t)$ directly by combining it linearly with the predicted type:
\begin{align}
P(\tau_t|c_l,c_r,y_t) 
=&\frac{P(y_t|\tau_t, c_l,c_r)}{2P(y_t|c_l,c_r)}\cdot P(\tau_t|c_l,c_r) \nonumber \\
 &+ \frac{P(c_l,c_r|\tau_t, y_t)}{2P(c_l,c_r|y_t)}\cdot P(\tau_t|y_t) \nonumber \\
=&\alpha \cdot P(\tau_t|c_l,c_r) \nonumber \\
 &+ \beta \cdot P(\tau_t|y_t) \label{prior_bayes_approx}
\end{align}

The coefficients $\alpha$ and $\beta$ are free parameters and are tuned on a small amount of withheld data. 

\subsubsection{Training with type priors}
\label{prior_as_penalty}
An alternative for incorporating the pre-computed $P(\tau_t|y_t)$ is to use it during training to regularize the type distribution. We use the following optimization criterion to compute the loss for each word:
\begin{align}
L =& H(P(y_{i}|c_l,c_r), P(\hat{y}_{i}|c_l,c_r)) \nonumber \\
 &+ \lambda\cdot||KL(P(\tau_{i}|c_l,c_r), P(\tau_{i}|y_{i}))||^2
\label{prior_penalty}
\end{align}
where $\hat{y}_{i}$ is the actual word, $H(.)$ is the cross entropy function, and $KL(.)$ measures the KL divergence between two distributions. A hyperparameter $\lambda$ (tuned on validation data) controls the relative contribution of the two loss terms. The new loss forces the learned type distribution, $P(\tau_{i}|c_l,c_r)$, to be close to the expected distribution $P(\tau_{i}|y_{i})$ given the information in the database. This loss is specifically tailored to help with unsupervised NER.

%% file: experiments.tex
\section{Experiments}
\label{exp}
We evaluate KALM on two tasks: language modeling and NER. We use two datasets: \emph{Recipe} used only for LM evaluation and \emph{CoNLL 2003} used for both the LM and NER evaluations.

\subsection{Data}
\textbf{Recipe}\hspace{0.5cm} The recipe dataset\footnote{Crawled from \url{http://www.ffts.com/recipes.htm}} is composed of $95,786$ recipes, 
We follow the same preprocessing steps as in \citet{parvez2018building} and divide the crawled dataset into training, validation and testing. A typical sentence after preprocessing looks like the following: \say{in a large mixing bowl combine the butter sugar and the egg yolks}. The entities in the recipe KB are recipe ingredients. The $8$ supported entity types are \textit{dairy}, \textit{drinks}, \textit{fruits}, \textit{grains}, \textit{proteins}, \textit{seasonings}, \textit{sides}, and \textit{vegetables}. In the sample sentence above, the entity names are \textit{butter}, \textit{sugar}, \textit{egg} and \textit{yolks}, typed as \textit{dairy}, \textit{seasonings}, \textit{proteins} and \textit{proteins}, respectively.

\textbf{CoNLL 2003}\hspace{0.5cm} Introduced in \citet{tjong2003introduction}, the CoNLL 2003 dataset is composed of news articles. It contains text and named entity labels in English, Spanish, German and Dutch. We experiment only with the English version. We follow the CoNLL labels and separate the KB into four entity types: \textit{LOC} (location), \textit{MISC} (miscellaneous), \textit{ORG} (organization), and \textit{PER} (person). 

Statistics about the recipe and the CoNLL 2003 dataset are presented in Table \ref{data_stats}.

\begin{table}[ht!]
\centering
\begin{tabular}{ c | c c c }
  & train & valid & test \\ \hline
 \#sent & 61302 & 15326 & 19158 \\  
 \#tok & 7223474 & 1814810 & 2267797   
\end{tabular}
\quad
\begin{tabular}{ c | c c c }
  & train & valid & test \\ \hline
 \#sent & 14986 & 3465 & 3683 \\  
 \#tok & 204566 & 51577 & 46665   
\end{tabular}
\caption{Statistics of recipe and CoNLL 2003 datasets}
\label{data_stats}
\end{table}

The information about the entities in each of the KBs is shown in Table \ref{kb_stats}. The recipe KB is provided along with the recipe dataset\footnote{The KB can be found in \url{https://github.com/uclanlp/NamedEntityLanguageModel}} as a conglomeration of typed ingredients. 
The KB used by CoNLL 2003 is extracted from WikiText-2. We filtered the entities which are not belonging to the 4 types of CoNLL 2003 task.
\begin{table}[ht!]
\small
\centering
\begin{tabular}{c | c c c c}
    type & dairy & drinks & fruits & grains \\\hline
    \#entities & 80 & 84 & 110 & 158 \\\hline \hline
    type & proteins & seasonings & sides & vegetables \\\hline
    \#entities & 316 & 180 & 140 & 156
\end{tabular}
\quad
\begin{tabular}{ c | c c c c}
    type & LOC & MISC & ORG & PER \\ \hline
    \#entity words & 1503 & 1211 & 3005 & 5404
\end{tabular}
\caption{Statistics of recipe and CoNLL 2003 KBs}
\label{kb_stats}
\end{table}

\subsection{Implementation details}
We implement KALM by extending the AWD-LSTM\footnote{\url{https://github.com/salesforce/awd-lstm-lm}} language model in the following ways:

\textbf{Vocabulary}\hspace{0.5cm} We use the entity words in Table \ref{kb_stats} to form $V_1,...,V_{K}$ We extract $51,677$ general words in the recipe dataset, and $17,907$ general words in CoNLL 2003 to form $V_0$. 
Identical words that fall under different entity types, such as \textit{Washington} in \textit{George Washington} and \textit{Washington D.C.}, share the same input embeddings.


\textbf{Model}\hspace{0.5cm} The model has an embedding layer of $400$ dimensions, LSTM cell and hidden states of $1,150$ dimensions, and $3$ stacked LSTM layers. We scale the final LSTM's hidden and cell states to $400$ dimensions, and share weights between the projection layer $\boldsymbol{W}^p$ and the word embedding layer. 
Each entity type in the knowledge base is represented by a trainable $100$-dimensional embedding vector.
When concatenating the weighted average of the type embeddings to the input, we expand the input dimension of the first LSTM layer to $500$. All trainable parameters are initialized uniformly randomly between $-0.1$ and $0.1$, except for the bias terms in the decoder linear layer, which are initialized to $0$.

For regularization, we adopt the techniques in AWD-LSTM, and use an LSTM weight dropout rate of $0$, an LSTM first-layers locked dropout rate of $0.3$, an LSTM last-layer locked dropout rate of $0.4$, an embedding Bernoulli dropout rate of $0.1$, and an embedding locked dropout rate of $0.65$. Also, we impose L2 penalty on LSTM pre-dropout weights with coefficient $1$, and L2 penalty on LSTM dropout weights with coefficient $2$, both added to the cross entropy loss.

\textbf{Optimization}\hspace{0.5cm} We use the same loss penalty, dropout schemes, and averaged SGD (ASGD) as in \citet{merity2017regularizing}. The initial ASGD learning rate is $10$, weight decay rate is $1.2\times 10^{-6}$, non-monotone trigger for ASGD is set to $5$, and gradient clipping happens at $0.25$. The models are trained until the validation set performance starts to decrease.

\subsection{Language modeling}
First, we test how good KALM is as a language model compared to two baselines.
\subsubsection{Baselines}
\begin{itemize}
\item  \textbf{AWD-LSTM} \citep{merity2017regularizing} is the state-of-the-art word-level language model as measured on WikiText-2 and Penn Treebank. It uses ASGD optimization, a new dropout scheme and novel penalty terms in the loss function to improve over vanilla LSTM LMs.

\item  \textbf{Named-entity LM} (NE-LM) \citep{parvez2018building} consists of a \textit{type model} that outputs $P(\tau_{i+1}|\tau_{i}, \tau_{i-1}, ...)$ and an \textit{entity composite model} that outputs $P(y_{i+1} | \{y_i, \tau_{i}\}, \{y_{i-1}, \tau_{i-1}\}, ...)$. The \textit{type model} is trained on corpora with entity type labels, whereas the \textit{entity composite model} has an input for words and another input for the corresponding types, and so needs to be trained on both the labeled corpus and the unlabeled version of the same corpus. At inference time, a joint inference heuristic aggregates type model and entity composite model predictions into a word prediction. Since both models require type labels as input, each generation step of NE-LM requires not only the previously generated words $[y_i, y_{i-1}, ...]$, but also the type labels for these words $[\tau_i, \tau_{i-1}, ...]$.
\end{itemize}

\subsubsection{Results}
For language modeling we report word prediction perplexity on the recipe dataset and CoNLL 2003. Perplexity is defined as the following.
\begin{align}
PP &= e^{-\sum\limits_{t=1}^{N}\frac{1}{N}\log P(y_t)} \nonumber \\
 &=\sqrt[N]{\prod\limits_{t=1}^{N}\frac{1}{P(y_t)}}
\label{perplexity}
\end{align}
We use publicly available implementations to produce the two baseline results. We also compare the language models in the bidirectional setting, which the reference implementations do not support. In that setting, we transform both models in NE-LM to be bidirectional.

\begin{table*}[ht!]
\centering
    \begin{tabular}{c c | c c | c c}
        & \multirow{2}{*}{model} & \multicolumn{2}{|c|}{unidirectional} & \multicolumn{2}{|c}{bidirectional} \\ \cline{3-6}
        & & validation & test & validation & test \\ \hline
        \multirow{3}{*}{Recipe} 
        & AWD-LSTM & 3.14 & 2.99 & 1.98 & 2 \\
        & NE-LM & 2.96 & 2.24 & 1.85 & 1.73 \\
        & KALM & \textbf{2.75} & \textbf{2.20} & \textbf{1.85} & \textbf{1.71} \\ \hline 
        \multirow{3}{*}{CoNLL 2003} 
        & AWD-LSTM & 5.48 & 5.94 & 4.85 & 5.3 \\
        & NE-LM & 5.67 & 5.77 & 4.68 & 4.94 \\
        & KALM & \textbf{5.36} & \textbf{5.43} & \textbf{4.64} & \textbf{4.74} \\ \hline 
    \end{tabular}
    \caption{Language modeling results on the recipe and CoNLL 2003 datasets}
    \label{lm_res}
\end{table*}



\textbf{Discussion}\hspace{0.5cm} Table \ref{lm_res} shows that KALM outperforms the two baselines in both unidirectional and bidirectional settings on both datasets. The improvement relative to NE-LM is larger in the unidirectional setting compared to the bidirectional setting. We conjecture that this is because in that setting NE-LM trains a bidirectional NER in a supervised way. The improvement relative to NE-LM is larger on CoNLL 2003 than on the recipe dataset. We believe that the inference heuristic used by NE-LM is tuned specifically to recipes and is less suitable to the CoNLL setting. 

We also find that training KALM on more unlabeled data further reduces the perplexity (see Table \ref{wikitext}), and study how the quality of the KB affects the perplexity. We discuss both these results in Section \ref{sec:ner}. 

\subsection{NER}
\label{sec:ner}
In this section, we evaluate KALM in NER against two supervised baselines.

\subsubsection{Baselines}
\label{supervised_ner}
We train two supervised models for NER on the CoNLL 2003 dataset: a \textbf{biLSTM} and a \textbf{CRF-biLSTM}. We replicate the hyperparameters used by \citet{lample2016neural}, who demonstrate the state-of-the-art performance on this dataset. We use a word-level model, and $100$ dimensional pretrained GloVe embeddings \citep{pennington2014glove} for initialization. We train for $50$ epochs, at which point the models converge.

\subsubsection{Results}
We evaluate the unsupervised KALM model under the following configurations: 
\begin{itemize}
\item Basic: bidirectional model with aggregated type embeddings fed to the input at the next time step;
\item With type priors: using $P(\tau_t|y_t)$ in the two ways described in Section \ref{use_prior};
\item Extra data: Since KALM is unsupervised, we can train it on extra data. We use the WikiText-2 corpus in addition to the original CoNLL training data.
\end{itemize}

WikiText-2 is a standard language modeling dataset released with \citet{merity2016pointer}. It contains Wikipedia articles from a wide range of topics. In contrast, the CoNLL 2003 corpus consists of news articles. Table \ref{wikitext} show statistics about the raw / characer level WikiText-2 and the CoNLL 2003 corpora. 
Despite the domain mismatch between the WikiText and CoNLL corpora, the WikiText coverage of the entity words that exist in the CoNLL dataset is high. Specifically, most of the person, location and organization entity words that appear in CoNLL either have a Wikipedia section, or are mentioned in a Wiki article. Therefore, we expect that the addition of WikiText can guide the unsupervised NER model to learn better entity type regularities.
Indeed, the result presented in the rightmost column of Table \ref{wikitext} shows that when adding 
WikiText-2 to CoNLL 2003, the perplexity for the KALM model for the news text of CoNLL 2003 is decreased: from $4.69$ down to $\textbf{2.29}$.

\begin{table*}[ht!]
\centering
\begin{tabular}{ c|c|c|c }
 entity & unique entity & size & LM \\ 
 ratio & ratio & ratio & perplexity \\ 
 \hline
92.80\% & 82.56\% & 2.62 & 2.29 : 4.69 
\end{tabular}
\caption{Characterization of WikiText-2 relative to CoNLL 2003 training set. \textit{Entities extracted from WikiText cover $92.80\%$ of the entities in CoNLL 2003 overall, and cover $82.56\%$ of the unique entities. WikiText's size is $2.62$ times as large. And adding WikiText to CoNLL training reduces the perplexity from $4.69$ to $2.29$.}}
\label{wikitext}
\end{table*}

We show NER results in Table \ref{ner_res}. The table lists the F1 score for each entity types, as well as the overall F1 score.

\begin{table*}[ht!]
\centering
\begin{tabular}{c|c|c|c|c|c|c}
\multicolumn{2}{c|}{}& LOC & MISC & ORG & PER & overall \\ \hline 
\multirow{4}{*}{unsupervised} & basic$^{0}$ & 0.75 & 0.67 & 0.64 & 0.83 & 0.72 \\ \cline{2-2}
& +$P(\tau|y)$ in dec$^{1}$ & 0.81 & 0.67 & 0.65 & 0.88 & 0.76 \\ \cline{2-2}
& +wiki-concat$^{2}$ & 0.83 & 0.83 & 0.76 & 0.95 & 0.84 \\ \cline{2-2}
& +wiki-concat+$P(\tau|y)$ in dec$^{3}$ & 0.84 & \textbf{0.84} & 0.77 & \textbf{0.96} & 0.86 \\ \hline \hline 
\multirow{2}{*}{supervised} & biLSTM & 0.88 & 0.71 & 0.81 & 0.95 & 0.86 \\
& CRF-biLSTM & \textbf{0.90} & 0.76 & \textbf{0.84} & 0.95 & \textbf{0.89} \\ \hline
\end{tabular}
\caption{Results of KALM (above the double line in the table) and supervised NER models (under the double line). Superscript annotations: \textit{
$0$: The basic bidirectional KALM with type embedding features as described in Section \ref{bidirectional}. 
$1$: Adding $P(\tau|y)$ in decoding, as described in Section \ref{prior_in_dec}, where $\alpha$ and $\beta$ are tuned to be $0.4$ and $0.6$. 
$2$: The basic model trained on CoNLL 2003 concatenated with WikiText-2.
$3$: Adding $(\tau|y)$ in decoding, with the model trained on CoNLL 2003 concatenated with WikiText-2.
}
}
\label{ner_res}
\end{table*}

\textbf{Discussion}\hspace{0.5cm} 
Even the basic KALM model learns context well -- it achieves an overall F1 score of $0.72$ for NER. This illustrates that KALM has learned to model entity classes entirely from surrounding context. Adding prior information as to whether a word represents different entity types helps to bring the F1 score to $0.76$. 

The strength of an unsupervised model is that it can be trained on large corpora. Adding the Wikitext-2 corpus improves the NER score of KALM to $0.84$.

\begin{figure}[ht!]
\centering
\begin{minipage}{.45\textwidth}
    \includegraphics[width=0.9\linewidth]{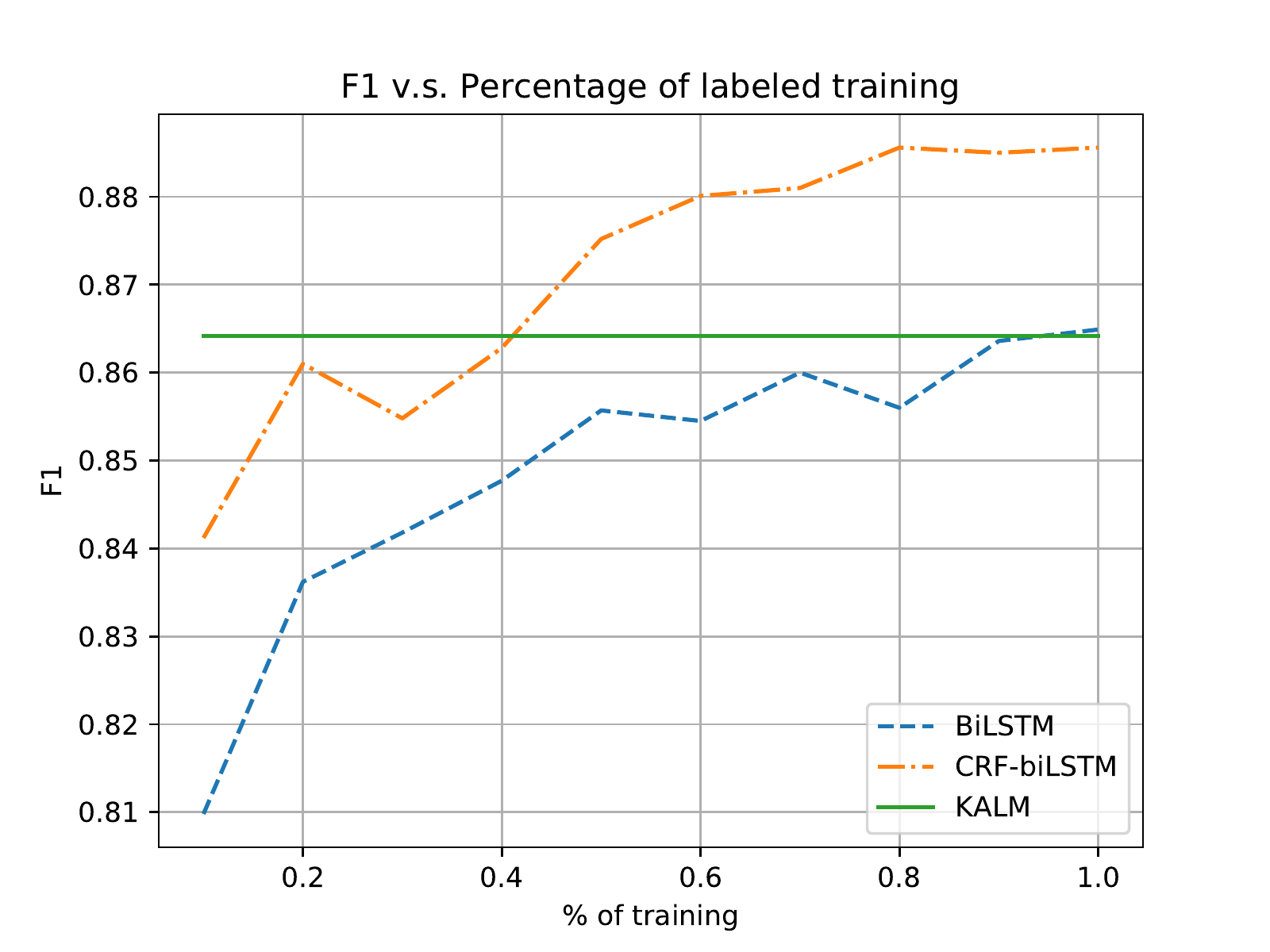}
    \captionof{figure}{Unsupervised v.s. supervised NER trained on different portions of training data}
    \label{fig:crfbilstm}
\end{minipage}%
\hfill
\begin{minipage}{.45\textwidth}
    \includegraphics[width=0.9\linewidth]{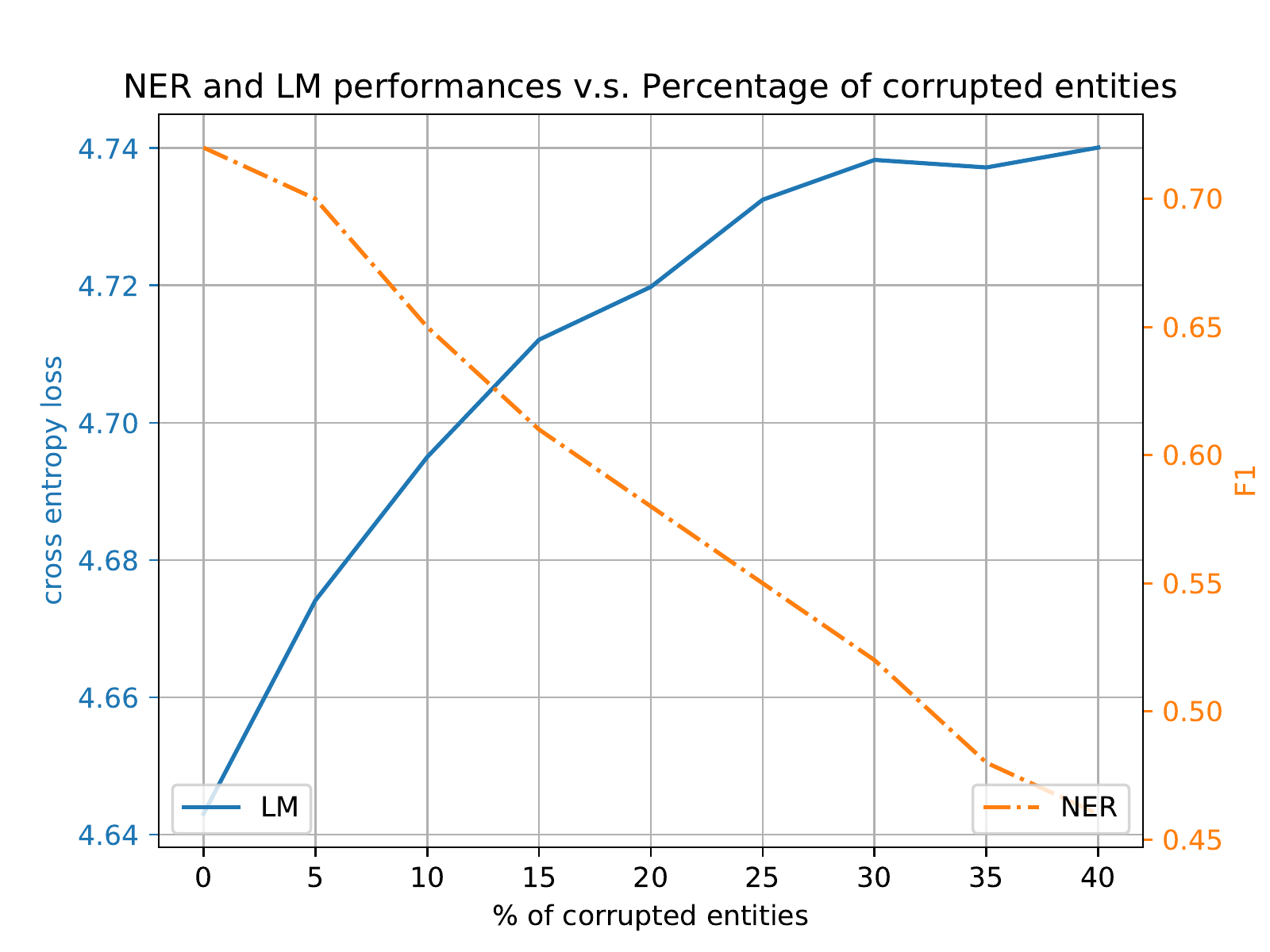}
    \captionof{figure}{NER and LM performances on different portions of mislabeled entities}
    \label{fig:reduce_ent}
\end{minipage}
\end{figure}

To give a sense of how the unsupervised models compare with the supervised model with respect to training data size, we trained biLSTM and CRF-biLSTM on a randomly sampled subset of the training data of successively decreasing sizes. The resulting F1 scores are shown in Figure \ref{fig:crfbilstm}.


Our best model 
scores $0.86$, same as a CRF-biLSTM trained on around $40\%$ of the training data. It is less than $0.03$ behind the best supervised CRF-biLSTM. 
The best KALM model almost always scores higher than biLSTM without the CRF loss.

Lastly, we perform an ablation experiment to gauge how sensitive KALM is to the quality of the knowledge base. Previous studies \cite{liu2016effective, zhang2012big} have shown that the amount of knowledge retrieved from KBs can impact the performance of NLP models such as relation extraction systems substantially.
In this experiment, we deliberately corrupt the entity vocabularies $V_0,...,V_{K-1}$ by moving a certain percentage of randomly selected entity words from $V_i$ to the general vocabulary $V_g$. Figure \ref{fig:reduce_ent} shows language modeling perplexities on the validation set, and NER F1 scores on the test set as a function of the corruption percentage. The language modeling performance stops reacting to KB corruption beyond a certain extent, whereas the NER performance keeps dropping as the number of entities removed from $V_1, V_2, ...$ increases. This result shows the importance of the quality of the KB to KALM.

%% file: conclusion.tex
\section{Conclusion}
\label{conclusion}
We propose Knowledge Augmented Language Model (KALM), which extends a traditional RNN LM with information from a Knowledge Base.
We show that real-world knowledge can be used successfully for natural language understanding by using a probabilistic extension. The latent type information is trained end-to-end using a predictive objective without any supervision. We show that the latent type information that the model learns can be used for a high-accuracy NER system. We believe that this modeling paradigm opens the door for end-to-end deep learning systems that can be enhanced with latent modeling capabilities and trained in a predictive manner end-to-end. In ways this is similar to the attention mechanism in machine translation where an alignment mechanism is added and trained latently against the overall translation perplexity objective. As with our NER tags, machine translation alignments are empirically observed to be of high quality.

In future work, we look to model other types of world knowledge beyond named entities using predictive learning and training on large corpora of text without additional information, and to make KALM more robust against corrupted entities.